\documentclass[10pt,letterpaper]{article}

\usepackage[margin=1in]{geometry}
\usepackage{times}
\usepackage[authoryear,round]{natbib}
\usepackage[utf8]{inputenc}
\usepackage[T1]{fontenc}
\usepackage{hyperref}
\usepackage{xurl}
\hypersetup{
  hypertexnames=false,
  hidelinks,
  pdfauthor={Xueting Fang, Zehui Li, Yang Yang, Camilla Giovino, Shubh K. Patel, Shailly Prajapati, Vallijah Subasri, Caihua Shan},
  pdftitle={KlinikeBench: Evaluating Language Models Beyond Diagnostic Accuracy},
  pdfsubject={Interactive clinical language-model evaluation}
}
\usepackage{booktabs}
\usepackage{tabularx}
\usepackage{float}
\usepackage{array}
\usepackage{amsfonts}
\usepackage{amsmath}
\usepackage{nicefrac}
\usepackage{microtype}
\usepackage{xcolor}
\usepackage{listings}
\usepackage{enumitem}
\usepackage{graphicx}
\usepackage{capt-of}
\usepackage{wrapfig}
\usepackage{needspace}
\newcommand{\finishwrap}{\par\ifnum\value{WF@wrappedlines}>1\vspace{\dimexpr\value{WF@wrappedlines}\baselineskip-\baselineskip\relax}\fi\WFclear}
\usepackage{tikz}
\usepackage{pgfplots}
\usepackage[capitalize,noabbrev]{cleveref}
\usetikzlibrary{positioning,arrows.meta,shapes.geometric,fit,backgrounds,calc}
\pgfplotsset{compat=1.16}

\definecolor{caaInk}{HTML}{2A2522}
\definecolor{caaInkDim}{HTML}{6E6259}
\definecolor{caaLine}{HTML}{D9D2C7}
\definecolor{caaPanel}{HTML}{FAF8F4}
\definecolor{caaSunken}{HTML}{F2EEE8}
\definecolor{caaAccent}{HTML}{0F8F8C}      % teal
\definecolor{caaAccentSoft}{HTML}{D5ECEB}
\definecolor{caaIndigo}{HTML}{2E5BC2}
\definecolor{caaIndigoSoft}{HTML}{D8E1F4}
\definecolor{caaCrimson}{HTML}{C0395A}
\definecolor{caaCrimsonSoft}{HTML}{F4DCE2}
\definecolor{caaAmber}{HTML}{C58A2C}
\definecolor{caaAmberSoft}{HTML}{F4E5C9}
\definecolor{caaGreen}{HTML}{4F8C5C}
\definecolor{caaGreenSoft}{HTML}{D8E9DC}

\tikzset{
  caa-node/.style={rectangle, rounded corners=2pt, draw=caaLine, fill=caaPanel,
                   text=caaInk, align=center, minimum height=8mm,
                   font=\sffamily\footnotesize, inner sep=4pt},
  caa-accent/.style={caa-node, draw=caaAccent, fill=caaAccentSoft, text=caaInk},
  caa-indigo/.style={caa-node, draw=caaIndigo, fill=caaIndigoSoft},
  caa-crimson/.style={caa-node, draw=caaCrimson, fill=caaCrimsonSoft},
  caa-amber/.style={caa-node, draw=caaAmber, fill=caaAmberSoft},
  caa-arrow/.style={->, >={Stealth[length=2.2mm]}, draw=caaInkDim, line width=0.5pt},
  caa-arrow-strong/.style={caa-arrow, draw=caaAccent, line width=0.7pt},
  caa-label/.style={font=\sffamily\scriptsize, text=caaInkDim},
}

\title{KlinikeBench: Evaluating Language Models Beyond Diagnostic Accuracy}

\author{%
  \normalsize Xueting Fang\textsuperscript{1}\thanks{Xueting Fang and Zehui Li contributed equally.}\quad
  Zehui Li\textsuperscript{2}\footnotemark[1]\thanks{Corresponding author: Zehui Li (\href{mailto:zl6222@ic.ac.uk}{\nolinkurl{zl6222@ic.ac.uk}}).}\quad
  Yang Yang\textsuperscript{3}\quad
  Camilla Giovino\textsuperscript{4}\\[3pt]
  \normalsize Shubh K. Patel\textsuperscript{4}\quad
  Shailly Prajapati\textsuperscript{4}\quad
  Vallijah Subasri\textsuperscript{5}\quad
  Caihua Shan\textsuperscript{6}\\[8pt]
  \small \textsuperscript{1}Zhejiang University\quad
  \textsuperscript{2}Imperial College London\quad
  \textsuperscript{3}Nanchang University\\[2pt]
  \small \textsuperscript{4}University of Toronto\quad
  \textsuperscript{5}Peter Munk Cardiac Centre, University Health Network\quad
  \textsuperscript{6}Microsoft
}
\date{}

\begin{document}

\maketitle

\begin{abstract}

Most clinical benchmarks evaluate language models (LMs) on diagnosis using complete case descriptions. In clinical practice, however, patients present information in different ways, and clinicians must obtain relevant history and determine which examinations are needed before reaching a diagnosis. Diagnostic accuracy alone therefore cannot establish whether an agent gathered essential information or conducted an appropriate clinical assessment. Furthermore, existing benchmarks lack professional clinicians' verification. To address this gap, we introduce KlinikeBench, a benchmark of 333 clinician-authored tasks, each providing an isolated sandbox environment with a virtual patient, clinical tools, and task-specific success criteria. More than 35 clinicians contributed to case authoring and benchmark evaluation. In an empirical study, clinicians gave simulated dialogues higher mean quality ratings than reference conversations, which is adapted from real conversation. In each task, an LM has a fixed budget of turns to communicate with the patient, ask about relevant history, request examinations, follow action constraints, and record a final diagnosis. We score these steps separately as well as together. Across 31 models and seven model families, the best-performing models (e.g., GPT-6-astra and Claude Opus 5) succeed on $<30\%$ of tasks, even though their diagnosis accuracy reaches 90.7\%. Some models benefit from talking with the patient; others diagnose well from a complete chart but perform much worse in conversation. Overall, KlinikeBench provides a testbed for evaluating the full clinical encounter and reveals a substantial gap between diagnostic accuracy and performance in interactive clinical assessment.

\end{abstract}

\begin{center}
\small Project website and resources: \url{https://zehui127.github.io/klinikebench/}
\end{center}

\begin{figure}[!t]
\centering
\includegraphics[width=\linewidth]{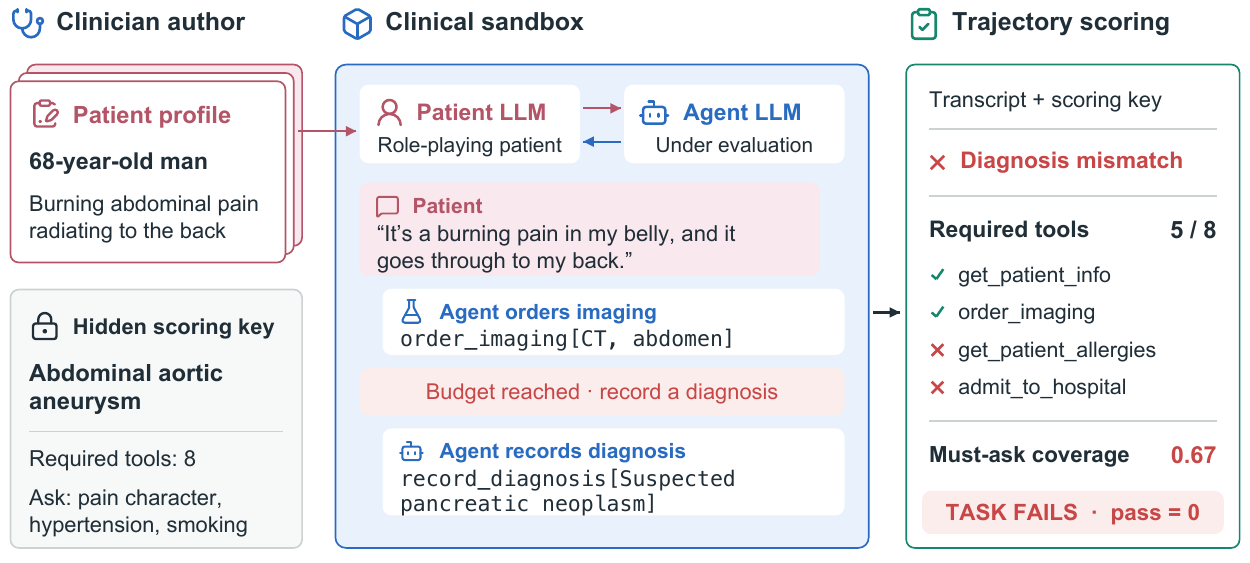}
\caption{How models are evaluated in KlinikeBench. A clinician authors the patient profile and hidden scoring key. The evaluated agent interacts with an LLM-simulated patient and executes clinical tools in a sandbox. The verifier scores the transcript for diagnosis, required tool use, action constraints, and must-ask coverage. This example fails because the diagnosis is incorrect and required tools and history topics are missed. Tool arguments, simulator messages, and tool checks are abbreviated.}
\label{fig:episode}
\end{figure}

\section{Introduction}\label{sec:intro}

Language models (LMs) are increasingly used to support clinical work, from summarizing records to answering patient questions and assisting with diagnostic decisions~\citep{futurereport2026,bedi2026medhelm}. Yet most established medical LM benchmarks test knowledge or diagnosis from a self-contained question or case description~\citep{medqa,medmcqa,pubmedqa,mmlu}. Such tasks provide the information needed to answer the question, leaving untested whether an agent can obtain that information through a clinical encounter. As clinical LM systems become more interactive, evaluation must account for how they gather evidence and carry out an assessment, as well as the conclusions they reach.

Recent clinical agent benchmarks move beyond static questions by exposing models to patient dialogue, tools, or electronic health records~\citep{agentclinic,medagentbench,2026PhysicianBench}. These environments make information gathering and sequential decisions part of the task. However, introducing interaction does not by itself establish whether an agent has completed an appropriate assessment. Two agents can reach the same correct diagnosis after very different encounters: one may ask about a crucial symptom and order the indicated test, while the other may guess early and omit both. Conversely, an incorrect diagnosis may follow missed history, incomplete clinical assessment, or faulty interpretation of the evidence. Distinguishing these failures requires explicit assessment of the questions asked and actions taken throughout the encounter. \Cref{tab:bench-compare} compares existing evaluations by their scoring, tools, data sources, and clinician verification. More detailed discussion on related work is in \Cref{app:related-work}.

\begin{table}[H]
\centering
\caption{Comparison with medical benchmarks and clinical-agent evaluations. KlinikeBench verification uses a survey of clinician judgments.}
\label{tab:bench-compare}
\small
\setlength{\tabcolsep}{3pt}
\renewcommand{\arraystretch}{1.04}
\begin{tabularx}{\linewidth}{@{}>{\raggedright\arraybackslash}p{.23\linewidth} >{\raggedright\arraybackslash}p{.115\linewidth} >{\raggedright\arraybackslash}p{.13\linewidth} >{\raggedright\arraybackslash}p{.105\linewidth} >{\raggedright\arraybackslash}p{.135\linewidth} >{\raggedright\arraybackslash}X@{}}
\toprule
\textbf{Benchmark} & \textbf{Metric} & \textbf{Interaction} & \textbf{Tools} & \textbf{Source} & \textbf{Clinician verification} \\
\midrule
MedQA \citeyearpar{medqa} & MCQA acc. & Single turn & None & Board exams & Exam keys \\
MMLU-med. \citeyearpar{mmlu} & MCQA acc. & Single turn & None & Exam material & Source answers \\
MedMCQA \citeyearpar{medmcqa} & MCQA acc. & Single turn & None & Entry exams & Expert questions \\
\midrule
HealthBench Pro. \citeyearpar{healthbench} & Rubric & Next reply & Varies & Clinical chats & $\geq$3 raters/item \\
AgentClinic \citeyearpar{agentclinic} & Diag. acc. & 20 turns & Tests + aids & Clinical cases & Dialogue ratings \\
MedAgentBench \citeyearpar{medagentbench} & Task success & Multi-step & FHIR APIs & EHRs & Task authoring \\
DeepRare \citeyearpar{deeprare} & Recall@$k$ & Ranked list & $>$40 & Rare diseases & Reference review \\
\midrule
\textbf{KlinikeBench (ours)} & \textbf{pass@1} & \textbf{32 turns} & \textbf{140} & \textbf{Clinicians} & \textbf{Clinician survey} \\
\bottomrule
\end{tabularx}
\end{table}

In clinical practice, patients present information incrementally and may not volunteer relevant history. Clinicians must decide what to ask, which examinations to request, and when they have enough evidence to make a diagnosis. Inspired by this workflow, we introduce \textbf{KlinikeBench}, a benchmark of 333 clinician-authored tasks that evaluate the full clinical encounter. As illustrated in \Cref{fig:episode}, each task combines a patient profile, an interactive clinical sandbox, and a hidden scoring key. An LLM role-plays the patient, while the evaluated agent elicits history, executes clinical tools, and records a diagnosis within a fixed interaction budget. Tool handlers retrieve information, update patient state, and serve case-specific clinical test results. The verifier assesses the resulting transcript against the clinician-authored requirements, including diagnostic correctness, required tool use, action constraints, and must-ask topics.

\begin{figure}[!ht]
\centering
\includegraphics[width=\linewidth]{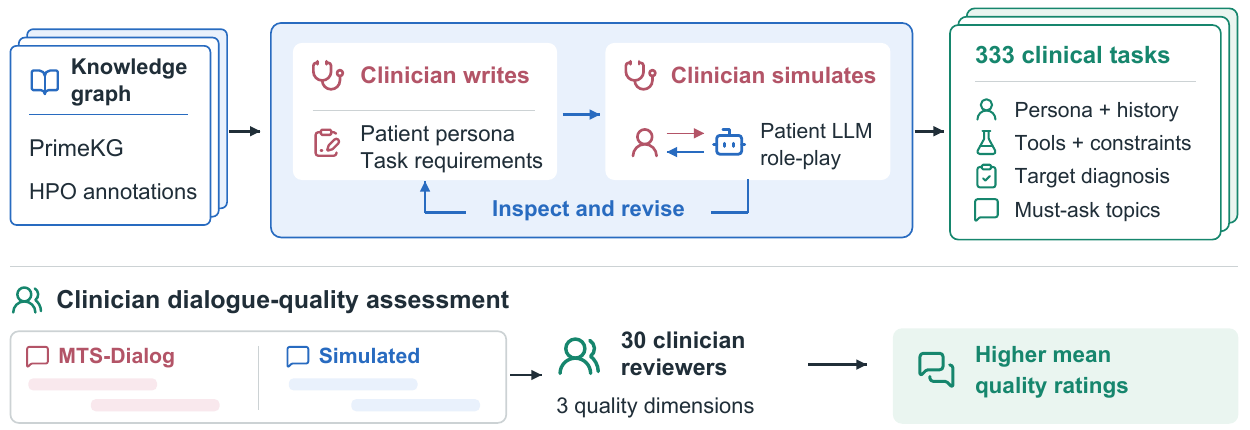}
\caption{Clinicians author and revise tasks through live simulation. In a separate assessment, virtual-patient dialogues receive higher mean quality ratings than references (shown in \Cref{tab:clinician-survey}).}
\label{fig:construction}
\label{fig:overview}
\end{figure}

To ensure the robustness of each task, KlinikeBench adopts clinician-led task construction. Furthermore, a data annotation platform is developed for clinicians to write patient personas, test the simulated encounters, and revise task drafts (\Cref{fig:construction}). In a separate dialogue-quality study, 30 clinicians gave virtual-patient conversations higher mean ratings than MTS-Dialog~\citep{mtsdialog2023}, a dataset of realistic patient-clinician conversation (\Cref{sec:human-results}), confirming the quality of authored tasks. In terms of scoring criteria, our evaluation makes individual omissions visible: in the example in \Cref{fig:episode}, the agent requests abdominal imaging but records an incorrect diagnosis, completes only five of eight required tool uses, and covers 67\% of the must-ask topics. Reporting these components separately identifies the deficiencies behind a failed encounter; requiring all criteria to pass measures whether the agent completes the task as a whole.

Our contributions are threefold. \textbf{1) A clinician-authored interactive benchmark.} KlinikeBench provides 333 tasks pairing simulated patients and executable clinical tools with clinician-specified assessment requirements, supported by a platform for authoring and revising encounters. \textbf{2) An encounter-level evaluation protocol.} Strict \texttt{pass@1} requires correct diagnosis, complete clinical actions, and full must-ask coverage; component scores expose omissions that diagnostic accuracy alone can hide. \textbf{3) An empirical study of clinical-agent performance.} We evaluate 31 models across seven families: diagnostic accuracy reaches 90.7\%, but the best strict \texttt{pass@1} is only 29.4\%. We further examine interaction budgets, single- versus multi-turn diagnosis, patient and judge substitutions, clinician ratings of patient-dialogue quality, and supervised fine-tuning and reinforcement learning of Qwen3-4B. Together, these evaluations generate more than 13,000 agent trajectories, providing a resource for studying and developing  LMs for disease diagnosis.

\paragraph{Public resources.}
The \href{https://zehui127.github.io/klinikebench/}{project website} provides an interactive leaderboard and benchmark instructions. We release the \href{https://hub.harborframework.com/datasets/klinikebench/klinikebench}{333-task benchmark on Harbor Hub}; \href{https://github.com/Zehui127/klinikebench-train}{SFT and GRPO training and evaluation code}, including task splits and environment setup; \href{https://huggingface.co/datasets/anon-caa-neurips/KLINIKEBENCH-Trajectories}{13,962 agent trajectories and SFT data}; the Qwen3-4B \href{https://huggingface.co/anon-caa-neurips/qwen3-4b-clinical-long-bench-sft}{SFT} and \href{https://huggingface.co/anon-caa-neurips/qwen3-4b-clinical-long-bench-grpo}{SFT + GRPO checkpoints}; and an \href{https://huggingface.co/spaces/anon-caa-neurips/klinikebench-demo}{interactive consultation demo}. The training repository also documents the \href{https://github.com/Zehui127/klinikebench-train#2-start-the-key-proxy}{host-side key proxy} used for patient-simulator and judge requests. Dataset configurations and benchmark-version information are documented in the trajectory dataset card.

\section{KlinikeBench}\label{sec:benchmark-schema}\label{sec:platform}\label{sec:method}

KlinikeBench is a benchmark of clinician-authored clinical encounters, each with a simulated patient, clinical tools, and evaluation criteria. The task is to question the patient, gather evidence using the tools, and record a correct diagnosis while satisfying the case-specific requirements (\Cref{fig:episode}).

\subsection{Benchmark construction}\label{sec:platform-generator}\label{sec:generator}

A group of clinicians\footnote{The authoring group comprises 35 clinicians from Canada and China, all with direct patient-care experience. Among the 30 questionnaire respondents, 19 reported more than five years of clinical practice.} authored the 333 tasks in KlinikeBench through the workflow shown in \Cref{fig:construction}. Our annotation platform supports autocompletion, case archiving, and live patient--clinician simulation for iterative revision (Appendix~\ref{app:annotation-platform}). The following formalizes the task definition and the task authoring process. Because this work does not involve designing a custom evaluation harness, we use the terms agent and LM interchangeably throughout the remainder of the paper.

\paragraph{Disease-grounded case authoring.} Clinicians select a target disease and associated findings, then construct the presentation from which the agent must infer the diagnosis. PrimeKG and HPO annotations supply disease--phenotype associations~\citep{primekg,hpoa}, supplemented by public case reports~\citep{mendeley}. For graph-derived cases, let $\mathcal{G}=(\mathcal{D}\cup\Phi,\mathcal{E})$ contain disease and phenotype nodes linked by \textsc{has\_phenotype} edges. The candidate findings for disease $d$ are
\begin{equation}
 \mathcal{N}_{\Phi}(d)=\{\phi\in\Phi:(d,\textsc{has\_phenotype},\phi)\in\mathcal{E}\}.
 \label{eq:phenotype-neighbourhood}
\end{equation}
An authored task is
\begin{equation}
 t=(d,\Phi^+,\Phi^-,\pi,\mathcal{R},M),
 \label{eq:authored-task}
\end{equation}
where $\Phi^+$ and $\Phi^-$ specify present and absent findings, $\pi$ is the patient persona, $\mathcal{R}$ contains tool-use requirements and action constraints, and $M$ contains must-ask topics. Clinicians select positive findings from $\mathcal{N}_{\Phi}(d)$, add case-specific details, and explicitly author negative findings with $\Phi^-\cap\Phi^+=\emptyset$; missing graph edges do not imply absent symptoms. They also specify accepted alternative diagnoses $A(d)$.

\paragraph{LLM assistance and clinician revision.} Clinicians write the persona and clinical requirements; the LLM assists with narrative refinement and structured formatting. Clinicians then inspect live patient--agent simulations and revise inconsistent responses, inappropriate disclosure, or infeasible requirements. Only approved tasks enter the released corpus $\mathcal{T}^{\star}$. The persona drives patient simulation, while the diagnosis and scoring criteria remain hidden from the evaluated agent.

\paragraph{Dialogue-quality assessment.}\label{sec:clinical-review} Thirty clinicians assessed 50 conversations for progressive history disclosure, emotional appropriateness, and human-likeness using five-point scales. We describe the review protocol and report the ratings in \Cref{sec:human-setup,sec:human-results}.

\subsection{Task format and evaluation}\label{sec:platform-schema}\label{sec:system}\label{sec:platform-criteria}\label{sec:platform-criteria-metrics}

\paragraph{Task and encounter workflow.} Each task combines a clinical briefing, patient simulator, initial clinical state, callable tools, and hidden scoring criteria (\Cref{fig:episode}). The agent receives the briefing and tool definitions, then questions the patient and gathers clinical information within 32 turns, including tool calls. The simulator responds from the authored persona and visible dialogue. Across the corpus, 140 tool names support record retrieval, state updates such as test orders and referrals, and access to cached, case-specific study findings. Missing information may yield unavailable responses or default values; Appendix~\ref{app:clinical-tools} details the tool interface and execution behavior. The agent ends the encounter by calling \texttt{record\_diagnosis}, or is prompted to do so when the turn limit is reached. The resulting dialogue and tool trace are passed to the verifier.

\paragraph{Data split.}\label{sec:data-split} We use all 333 tasks for the main model evaluation. To facilitate the usage to evaluate training algorithms in rare cases, we also provide a split of 283 training tasks and 50 held-out test tasks, with performance reported on the test set. 

\paragraph{Evaluation metrics.} For task $i$, the binary diagnosis gate $D_i$ checks whether the final diagnosis matches the reference or an accepted alternative. The binary required-tool gate $W_i$ passes only when all required tool calls are completed, and $A_i$ checks argument-level action constraints. Must-ask coverage $H_i$ is the fraction of must-ask topics explicitly raised by the agent, assessed by a fixed LLM judge. \textbf{Strict \texttt{pass@1}} is the fraction of tasks passing all four criteria with one rollout per task:
\begin{equation}
 s_i = D_i\,W_i\,A_i\,\mathbf{1}[H_i=1],
 \qquad
 \mathrm{pass@1} = \frac{1}{N}\sum_{i=1}^{N}s_i,
 \label{eq:task-success}
\end{equation}
where $N$ is the number of tasks. We also report \textbf{diagnosis accuracy}, $N^{-1}\sum_i D_i$; \textbf{required-tool completion}, $N^{-1}\sum_i W_i$; and \textbf{must-ask coverage}, $N^{-1}\sum_i H_i$. All metrics are percentages: required-tool completion counts tasks with all required tools completed, while must-ask coverage averages topic coverage per task. Strict \texttt{pass@1} requires all gates and all must-ask topics.

\subsection{Dataset statistics}\label{sec:corpus}\label{sec:dataset}\label{sec:corpus-overview}\label{sec:corpus-distributions}\label{sec:corpus-complexity}

\begin{figure}[!ht]
\centering
\begin{minipage}[t]{0.49\linewidth}
\vspace{0pt}
\centering
\begin{minipage}[c][146pt][c]{\linewidth}
\includegraphics[width=\linewidth]{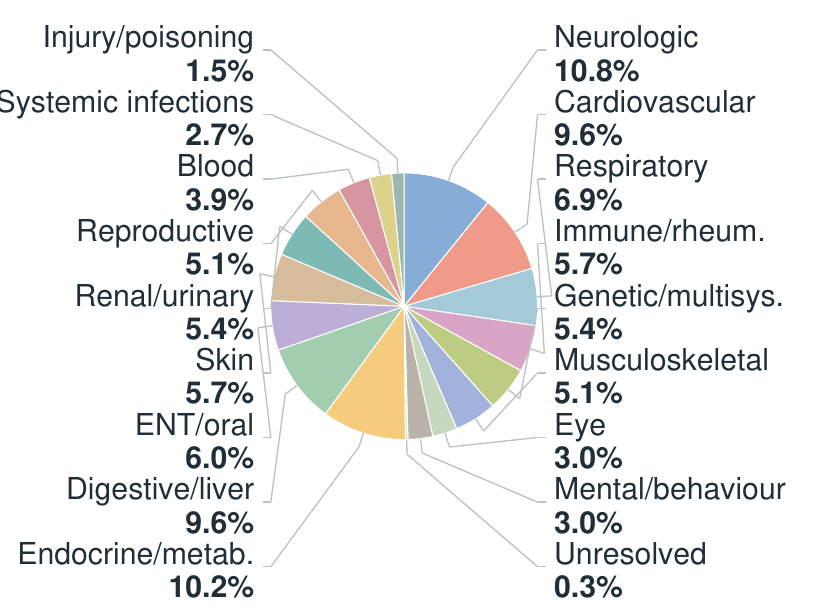}
\end{minipage}
\caption{Distribution of diseases spanning 17 clinical subcategories.}
\label{fig:dataset-statistics}
\label{fig:clinical-specialties}
\end{minipage}\hfill
\begin{minipage}[t]{0.49\linewidth}
\vspace{0pt}
\centering
\begin{minipage}[c][146pt][c]{\linewidth}
\includegraphics[width=\linewidth]{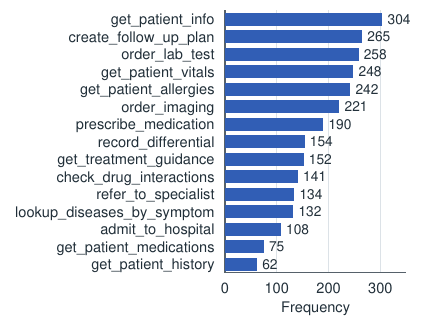}
\end{minipage}
\caption{Top 15 tools by task frequency, excluding final diagnosis.}
\label{fig:corpus-stats}
\end{minipage}
\end{figure}

\paragraph{Disease distribution.} \Cref{fig:dataset-statistics} shows disease groups; Appendix~\ref{app:corpus-statistics} reports corpus totals and interaction requirements. The corpus contains 304 distinct target-diagnosis labels: 278 occur in one task, 24 occur in two tasks, one occurs in three tasks, and one occurs in four. For visualization, we assign each task to one descriptive clinical group using its target diagnosis. The 226 cases are labeled as \emph{Internal Medicine} include cardiovascular,  digestive or liver, endocrine or metabolic, and  respiratory cases.

\paragraph{Evaluation Criteria.} The tasks require both history-taking and clinical actions: the median case specifies four must-ask topics and five required tools. \Cref{tab:dataset-statistics} summarizes these requirements and exposed tools. \Cref{fig:corpus-stats} shows the 15 most frequent tools, counted once per task across required and optional tools, excluding final diagnosis recording.

% Two vector figures; numerical corpus statistics are in the appendix.

\section{Experimental setup}\label{sec:experiments}\label{sec:bench-setup}

\subsection{Implementation details}\label{sec:eval}\label{sec:platform-harness}

We integrate KlinikeBench into Harbor~\citep{harbor}, packaging each case as a containerized task with clinical tools and a verifier. The evaluated model uses the ReAct-style \texttt{Terminus-2} harness to question the patient, execute tools through the \texttt{clinic} command interface, and record a diagnosis. The patient persona and scoring key remain inaccessible to the agent. Within the Harbor evaluation, each model receives the same tasks and tool interface, with one rollout per model--task pair.

\subsection{Models}

We evaluate 31 models from seven families: Claude, GPT, Gemini, Grok, DeepSeek, Kimi, and GLM (\Cref{tab:gates-results-full}). Medium reasoning effort is used across all models.

\subsection{Patient simulator and evaluators}\label{sec:evaluators}

As shown in \Cref{fig:sensitivity-ranks}, the ranking of models are relatively stable with different models as the patient simulator and judges. \texttt{GPT-5.4-mini} serves as both the patient simulator and history-topic judge in the main evaluation, with separate prompts and contexts. The patient responds from the clinician-authored persona and encounter history. The verifier checks diagnosis, required tool use, and action constraints against the hidden scoring criteria. The history judge evaluates each must-ask topic from the dialogue and returns a binary decision: the agent must explicitly ask about the topic; information volunteered by the patient alone does not count. Patient simulation uses temperature 0.7 and history judging uses temperature 0. We report strict \texttt{pass@1} and the component scores defined in \Cref{eq:task-success}.

\subsection{Training setup}\label{sec:training-setup}

We train Qwen3-4B~\citep{qwen3} using the benchmark's Harbor environment and \texttt{Terminus-2} harness, with 283 training tasks and 50 held-out test tasks. We first apply supervised fine-tuning on 1,235 successful frontier-model trajectories from the train split, then compare two GRPO runs~\citep{grpo} initialized from the same Supervised FineTuning (SFT) checkpoint: one optimizes diagnosis correctness, and the other a weighted reward combining all four gates. Both runs use identical data order and hyperparameters. Appendix~\ref{app:training-details} provides the data preparation, optimization settings, and evaluation schedule.

\subsection{Robustness and validation setup}\label{sec:robustness-setup}\label{sec:human-setup}

\paragraph{Judge and patient-simulator sensitivity.}
We assess how the choice of judge and patient simulator affects results on a 50-task subset using eight clinician models: Claude Opus 5, Claude Opus 5.5, Claude Sonnet 5, GPT-6-astra, GPT-5.6-sol, Grok-4.7, Gemini-3.8-flash, and Gemini-3.7-flash. We compare GPT-5.4-mini, GPT-5.4, Gemini-3.5-flash, and Grok-4.5 as judges, and GPT-5.4-mini, GPT-6-astra, Claude Sonnet 5, and Claude Haiku 4.5 as patient simulators. \Cref{fig:sensitivity-ranks} shows how the clinician-model rankings vary across these choices.

\paragraph{Interaction-budget sensitivity.} While the main evaluations use a limit of 32 turns, the sensitivity of the evaluation results on this limit needs to be studied. We evaluate GPT-6-astra, Claude-Opus-5, and Sonnet-5 on a 50-task subset under the default task-dependent step cap, caps of 45 and 60, and no step cap. The step budget includes dialogue and tool calls. With GPT-5.4-mini as patient and judge, we compare strict \texttt{pass@1} and component scores on matched tasks (\Cref{tab:budget-ablation}).

\paragraph{Clinician assessment.} To assess the quality of virtual patients, thirty clinicians each rated 50 conversations: 35 conversations between our virtual patient and a clinician and 15 reference conversations from MTS-Dialog~\citep{mtsdialog2023}.  MTS-Dialog is a collection of real patient-clinician conversations adopted from real patient records. The 50 conversations are mixed without labels; and then the clinicians are required to assess each conversation from three dimensions: progressive history disclosure, emotional appropriateness, and human-likeness. Each dimension was scored on a 5-point Likert scale (1 = completely unrealistic, 5 = indistinguishable from real patients), yielding 1,500 ratings per dimension and 4,500 ratings in total.

\begingroup
\setlength{\intextsep}{6pt}
\setlength{\columnsep}{10pt}

\section{Results}\label{sec:results}\label{sec:zeroshot}

We evaluate 31 models across seven families (\Cref{tab:gates-results}; full results in \Cref{tab:gates-results-full}). Models struggle to fully solve these tasks: Claude Opus 5 performs best at only 29.4\% strict \texttt{pass@1}. We then examine interaction budgets, dialogue, evaluator sensitivity, patient quality, and small-model training.

\begin{table}[!t]
\centering
\caption{Performance of 18 models on 333 tasks; all 31 appear in \Cref{tab:gates-results-full}. Scores are percentages: diagnosis accuracy, required-tool completion (Tools), and must-ask coverage. Strict \texttt{pass@1} requires all criteria (\Cref{sec:platform-criteria-metrics}). \textbf{Bold}: overall best; \underline{underlining}: best within families with multiple models. Citations link model reports or cards.}
\label{tab:gates-results}
\small
\renewcommand{\arraystretch}{0.90}
\setlength{\aboverulesep}{1.5pt}
\setlength{\belowrulesep}{1.5pt}
\setlength{\tabcolsep}{2pt}
\begin{tabular*}{\linewidth}{@{\extracolsep{\fill}}lrrrr@{}}
\toprule
Model & Strict \texttt{pass@1} & Diagnosis & Tools & Must-ask \\
\midrule
Claude Opus 5~\citep{claudeopus5} & \textbf{\underline{29.4\%}} & \textbf{\underline{90.7\%}} & 44.1\% & \underline{89.9\%} \\
Claude Opus 5.5~\citep{claudeopus55} & 28.8\% & \textbf{\underline{90.7\%}} & \underline{49.8\%} & 84.4\% \\
Claude Opus 4.8~\citep{claudeopus48} & 16.8\% & 89.2\% & 27.9\% & 85.5\% \\
Claude Sonnet 5~\citep{claudesonnet5} & 13.8\% & 87.4\% & 24.0\% & 85.1\% \\
Claude Haiku 4.5~\citep{claudehaiku45} & 2.7\% & 72.1\% & 8.1\% & 70.6\% \\
\midrule
GPT-6-astra~\citep{gpt6astra} & \underline{24.3\%} & \underline{87.4\%} & 39.3\% & 88.0\% \\
GPT-5.6-sol~\citep{gpt56} & 23.7\% & 86.2\% & 36.6\% & \textbf{\underline{90.8\%}} \\
GPT-5.4~\citep{gpt54} & 23.4\% & 80.2\% & \underline{53.2\%} & 73.5\% \\
GPT-4o~\citep{gpt4o} & 3.9\% & 66.4\% & 12.9\% & 79.5\% \\
\midrule
Gemini-3.8-flash~\citep{gemini38flash} & \underline{18.9\%} & \underline{80.8\%} & \textbf{\underline{53.8\%}} & 76.5\% \\
Gemini-3.5-flash~\citep{gemini35flash} & 15.6\% & 75.7\% & 43.8\% & \underline{76.6\%} \\
\midrule
Grok-4.7~\citep{grok47} & \underline{22.2\%} & \underline{83.2\%} & 35.1\% & 87.5\% \\
Grok-4.6~\citep{grok46} & 21.9\% & 81.4\% & 36.9\% & \underline{89.0\%} \\
Grok-4.3 & 21.3\% & 77.8\% & \underline{45.6\%} & 83.3\% \\
\midrule
DeepSeek-v4-pro~\citep{deepseekv4} & \underline{8.1\%} & \underline{29.4\%} & \underline{13.8\%} & \underline{87.8\%} \\
DeepSeek-v4-flash~\citep{deepseekv4} & 3.9\% & 18.0\% & 9.3\% & 81.3\% \\
\midrule
Kimi-K2.6~\citep{kimik26} & 23.4\% & 77.2\% & 48.7\% & 84.7\% \\
\midrule
GLM-5.1~\citep{glm5} & 4.8\% & 24.9\% & 10.8\% & 30.8\% \\
\bottomrule
\end{tabular*}
\end{table}

\paragraph{Correct diagnosis does not imply a complete assessment.}
Claude Opus 5 achieves 90.7\% diagnosis accuracy but only 29.4\% strict \texttt{pass@1} (\Cref{tab:gates-results}). Across all 31 models, mean diagnosis accuracy is 73.6\%, compared with 13.4\% strict success and 27.9\% required-tool completion. GPT-5.3-codex illustrates this gap: it diagnoses 84.4\% of cases correctly, yet completes the required tools on only 8.7\% and passes 3.6\% overall (\Cref{tab:gates-results-full}). Different models lead on different components: Gemini-3.8-flash has the highest required-tool completion (53.8\%), while GPT-5.6-sol has the highest must-ask coverage (90.8\%). Reaching the correct diagnosis therefore does not establish that the necessary history and clinical actions were completed. Appendix~\ref{app:additional-results} provides the cost--performance comparison.

\Needspace{150pt}
\begin{wrapfigure}{r}{0.45\textwidth}
\centering
\includegraphics[width=\linewidth]{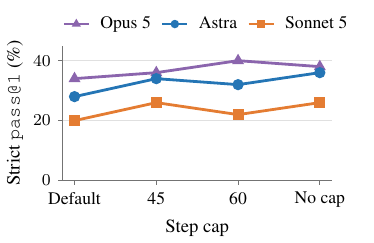}
\end{wrapfigure}
\paragraph{Pass rates do not improve monotonically with more steps.}\label{sec:budget-results}
We examine how the budget of interaction steps, which counts dialogue and tool calls, affects strict success while holding the ten-patient-turn
limit fixed.  Across the default (32-step), 45-step, 60-step, and uncapped conditions, strict pass@1 is 28/34/32/36\%
for GPT-6-astra, 34/36/40/38\% for Claude Opus 5,
and 20/26/22/26\% for Claude Sonnet 5. Overall, increasing the length of conversation does not consistently increase success, but there is positive influence on the outcome. 

\finishwrap
\Needspace{155pt}
\begin{wrapfigure}{r}{0.49\textwidth}
\centering
\includegraphics[width=\linewidth]{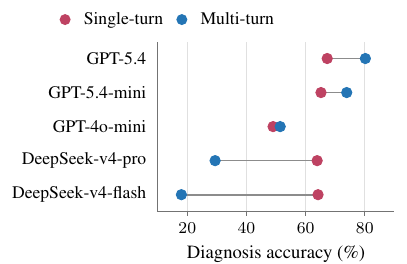}
\end{wrapfigure}
\paragraph{Single-turn vs. multi-turn diagnosis.}
We compare interactive diagnosis with a single-turn variant that supplies the full intake snapshot, including the chief complaint, demographics, severity, duration, history, medications, allergies, and family history. Both settings score the final diagnosis using the same reference-matching criterion. Across all 333 tasks, GPT-5.4, GPT-5.4-mini, and GPT-4o-mini improve by 12.9, 8.7, and 2.4 percentage points in the interactive setting. DeepSeek-v4-pro and DeepSeek-v4-flash instead decline by 34.5 and 46.2 points. Their similar single-turn scores (63.9\% and 64.2\%) become 29.4\% and 18.0\% in conversation. This comparison tests evidence gathering and dialogue control as well as diagnosis, while also changing how information is presented.

\finishwrap

\subsection{Ablations and patient-quality assessment}\label{sec:judge-results}\label{sec:human-results}

\paragraph{Judge substitutions broadly preserve rankings, but can change the leader.}
Re-scoring fixed trajectories with different must-ask judges broadly preserves clinician-model rankings (\Cref{fig:sensitivity-ranks}a). GPT-6-astra remains first, sometimes tied with Claude Opus 5.5, except under Gemini-3.5-flash, which ranks Opus 5.5 first. Broad ordering is therefore more stable than the identity of the leader. Only must-ask coverage is re-evaluated; the other gates remain fixed.

\paragraph{Patient simulation shifts scores but broadly preserves rankings.}
With GPT-5.4-mini as the patient simulator, GPT-6-astra ranks first. With Claude Haiku 4.5, Claude Opus 5 and Opus 5.5 share the top rank. Despite this change at the top, the overall ordering remains broadly similar. Because patient substitutions require fresh encounters, rank changes may reflect both the simulator choice and rollout variability. \Cref{fig:sensitivity-ranks}b shows the rankings, and \Cref{tab:full-sensitivity} reports the available scores.

\begin{figure}[!ht]
\centering
\includegraphics[width=\linewidth]{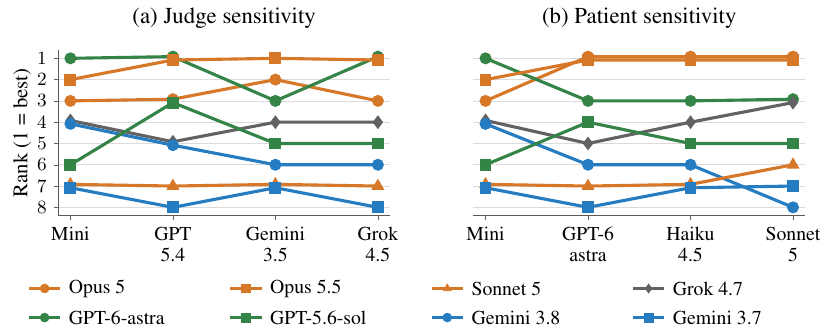}
\caption{Each point ranks a clinician by strict \texttt{pass@1} over 50 tasks under the indicated judge (a) or patient (b). Mini: GPT-5.4-mini; Gemini: Flash variants. Ties are slightly offset; full scores: \Cref{tab:full-sensitivity}.}
\label{fig:sensitivity-ranks}
\end{figure}

\Needspace{100pt}
\paragraph{Clinicians rate virtual-patient conversations favorably.}
Thirty clinicians assessed 35 virtual-patient conversations and 15 MTS-Dialog references (\Cref{tab:clinician-survey}). Virtual-patient mean ratings were 3.69 for history disclosure, 3.50 for emotional appropriateness, and 3.76 for human-likeness, compared with 3.43, 3.31, and 3.45 for the references. Ratings of four or five accounted for 53.2--64.9\% of virtual-patient ratings. Emotional appropriateness received the lowest score in both groups. These findings support the ability of our clinician-authored task profiles to produce realistic patient conversations.

\begin{table}[!ht]
\centering
\caption{Clinician ratings by conversation source (mean $\pm$ SD, scale 1--5). Thirty clinicians provide 1,050 ratings per dimension for 35 virtual-patient conversations and 450 for 15 MTS-Dialog references~\citep{mtsdialog2023}. SD summarizes individual ratings.}
\label{tab:clinician-survey}
\small
\setlength{\tabcolsep}{5pt}
\begin{tabular*}{\linewidth}{@{\extracolsep{\fill}}lrr@{}}
\toprule
Dimension & Virtual patient & MTS-Dialog \\
\midrule
Progressive history disclosure & $3.69\pm0.96$ & $3.43\pm1.05$ \\
Emotional appropriateness & $3.50\pm0.96$ & $3.31\pm1.08$ \\
Human-likeness & $3.76\pm1.07$ & $3.45\pm1.19$ \\
\bottomrule
\end{tabular*}
\end{table}

\Needspace{72pt}
\subsection{Supervised fine-tuning and reinforcement learning}\label{sec:training-results}

Standalone sandbox with Harbor enables the scalable RL training. We hereby evaluate Qwen3-4B on 50 held-out tasks after SFT and GRPO with diagnosis-only or multi-gate rewards (\Cref{eq:training-rewards}). Both RL runs share the SFT initialization and use Claude Opus 4.6 as patient and judge. We report the first 30 steps; \Cref{fig:qwen-rl-gates-extended} in the Appendix shows extended results and training details. 

\Needspace{195pt}
\begin{wrapfigure}{r}{0.63\textwidth}
\centering
\includegraphics[width=\linewidth]{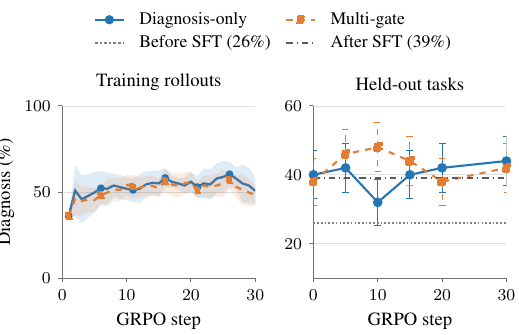}
\caption{Qwen3-4B diagnosis during GRPO. Left: five-step training means with $\pm$SD bands. Right: held-out accuracy with $\pm$SE bars (50 tasks).}
\label{fig:qwen-rl}
\end{wrapfigure}
\paragraph{Successful trajectories improve assessment beyond diagnosis.}
SFT raises strict \texttt{pass@1} from 0\% to 8\%, diagnosis accuracy from 26\% to 39\%, required-tool completion from 3\% to 33\%, and must-ask coverage from 39.7\% to 77.5\% (Appendix~\ref{app:additional-results}, \Cref{tab:qwen-results}). The strict \texttt{pass@1} gain is 8 percentage points, with a 95\% confidence interval from 2 to 15 percentage points. The intervals for the other three gains also lie above zero, supporting improvements in diagnosis, history-taking, and tool use.

\paragraph{RL gains remain uncertain.} We found three targeted metrics have different levels of difficulty to optimise. While SFT improves the model's perform across three metrics. RL could improve diagnosis, but struggles to improve the must-ask coverage. 
\Cref{fig:qwen-rl} separates training diagnosis rates from held-out performance. At step 30, diagnosis-only and multi-gate RL reach 44\% and 42\% diagnosis accuracy, respectively, compared with 39\% after SFT. Their strict \texttt{pass@1} scores are 10\% and 6\%, compared with 8\% after SFT; required-tool completion is 32\% in both runs. This is 2 percentage points above SFT for diagnosis-only RL and 2 points below it for multi-gate RL. However, both 95\% confidence intervals include possible gains and losses relative to SFT (Appendix~\ref{app:additional-results}). These early results do not establish that RL improves full-encounter success or that either reward is better.

\finishwrap

\Needspace{72pt}
\section{Qualitative analysis and Discussions}\label{sec:qualitative}

Overall, KlinikeBench serves as a strong proxy for evaluating LMs' capability in clinical settings,  requiring not only correct diagnosis, but proper dialogue and clinical tool usage. Across 31 models, diagnostic accuracy reaches 90.7\%, but strict success remains below 30\%; component scores expose incomplete history-taking and clinical actions. Patient and judge substitutions show that evaluation choices have limited effects on rankings. Clinician ratings provide descriptive evidence of dialogue quality, and Qwen3-4B fine-tuning improves assessment, while early RL gains remain uncertain, motivating richer evaluation alongside explicit completion criteria. Despite the overall robustness, there are still challenges, motivating the future research on model alignment and benchmark design.

\paragraph{Passing the gates does not guarantee a coherent encounter.} 
\Cref{fig:qualitative-vt} compares GPT-6-astra and Qwen3-4B after SFT and GRPO on the same held-out ventricular-tachycardia case, with Claude Opus 4.6 as the patient. Both trajectories pass the diagnosis and required-tool gates, cover all five must-ask topics, and receive a passing outcome. Their dialogue quality nevertheless differs: GPT-6-astra asks focused questions, explains its admission recommendation, and voluntarily records a specific diagnosis with low expressed confidence. Qwen covers the required topics but later produces garbled and incoherent statements, then records a broad diagnosis with high expressed confidence when forced by the step cap. This case shows that passing the gates does not establish dialogue coherence, diagnostic precision, or appropriate confidence.

\begin{figure}[!ht]
\centering
\includegraphics[width=\linewidth]{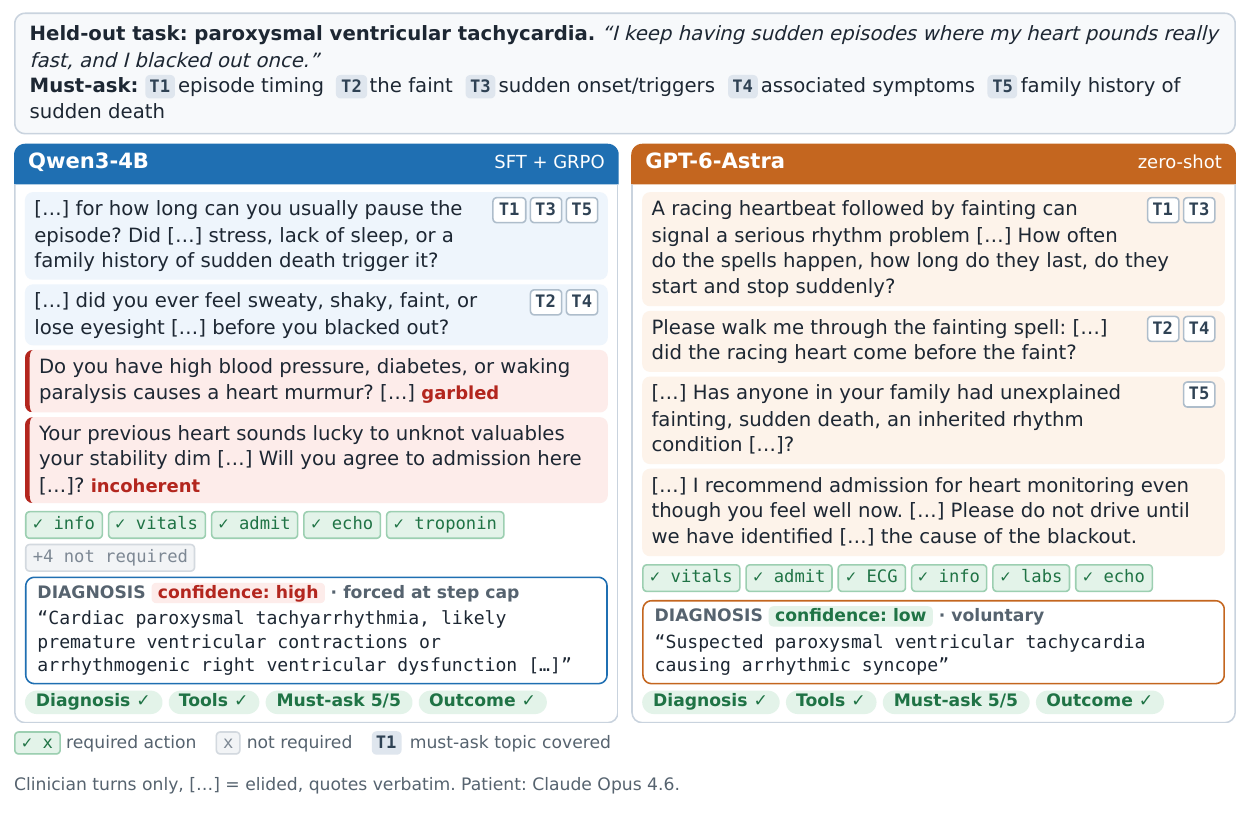}
\caption{Two passing trajectories on the same held-out case. Qwen3-4B (SFT + GRPO) and GPT-6-astra (zero-shot) satisfy the reported gates but differ in dialogue coherence and diagnosis recording. Excerpts are verbatim with omissions; the patient is Claude Opus 4.6.}
\label{fig:qualitative-vt}
\end{figure}
\endgroup

% \section{Conclusion}\label{sec:conclusion}\label{sec:discussion}

% Overall, KlinikeBench evaluates clinician-authored encounters requiring patient dialogue, clinical tools, and diagnosis. Across 31 models, diagnostic accuracy reaches 90.7\%, but strict success remains below 30\%; component scores expose incomplete history-taking and clinical actions. Patient and judge substitutions show that evaluation choices have limited effects on rankings. Clinician ratings provide descriptive evidence of dialogue quality, and Qwen3-4B fine-tuning improves assessment, while early RL gains remain uncertain. , motivating richer evaluation alongside explicit completion criteria.

\clearpage
\bibliographystyle{plainnat}
\bibliography{references}
\clearpage
\appendix

% A compact, appendix-only contents page with links to the actual headings.
\clearpage
\pdfbookmark[1]{Appendix contents}{appendix-contents}
\section*{Appendix contents}

\newcommand{\appendixcontentssection}[1]{%
  \par\addvspace{0.55\baselineskip}%
  \noindent\hyperref[#1]{\ref*{#1}\enspace\nameref*{#1}}%
  \dotfill\hyperref[#1]{\pageref*{#1}}\par
}
\newcommand{\appendixcontentssubsection}[1]{%
  \noindent\hspace*{1.5em}%
  \hyperref[#1]{\ref*{#1}\enspace\nameref*{#1}}%
  \dotfill\hyperref[#1]{\pageref*{#1}}\par
}

\appendixcontentssection{app:annotation-platform}
\appendixcontentssubsection{app:case-authoring}
\appendixcontentssubsection{app:live-simulation}
\appendixcontentssubsection{app:corpus-statistics}
\appendixcontentssection{app:clinical-tools}
\appendixcontentssubsection{app:tool-interface}
\appendixcontentssubsection{app:tool-state-updates}
\appendixcontentssubsection{app:tool-traces}
\appendixcontentssection{app:training-details}
\appendixcontentssection{app:additional-results}
\appendixcontentssubsection{app:complete-model-results}
\appendixcontentssubsection{app:cost-performance}
\appendixcontentssubsection{app:interaction-budget-scores}
\appendixcontentssubsection{app:full-sensitivity-results}
\appendixcontentssubsection{app:qwen-heldout-results}
\appendixcontentssubsection{app:grpo-gate-trajectories}
\appendixcontentssection{app:related-work}

\clearpage

\section{Clinician annotation platform}\label{app:annotation-platform}\label{app:workflow}

The annotation platform supports the construction and revision of the interactive cases described in \Cref{sec:platform-generator}. It brings disease selection, task editing, live simulation, and trajectory review into a shared workspace. Clinicians author the patient persona and clinical requirements, inspect simulated encounters, and decide which changes are needed before approving a case. The LLM assists with drafting and structured formatting. \Cref{fig:platform-authoring,fig:platform-simulation} show the authoring and simulation interfaces.

\subsection{Case authoring and LLM assistance}\label{app:case-authoring}

\begin{figure}[H]
\centering
\includegraphics[width=\linewidth]{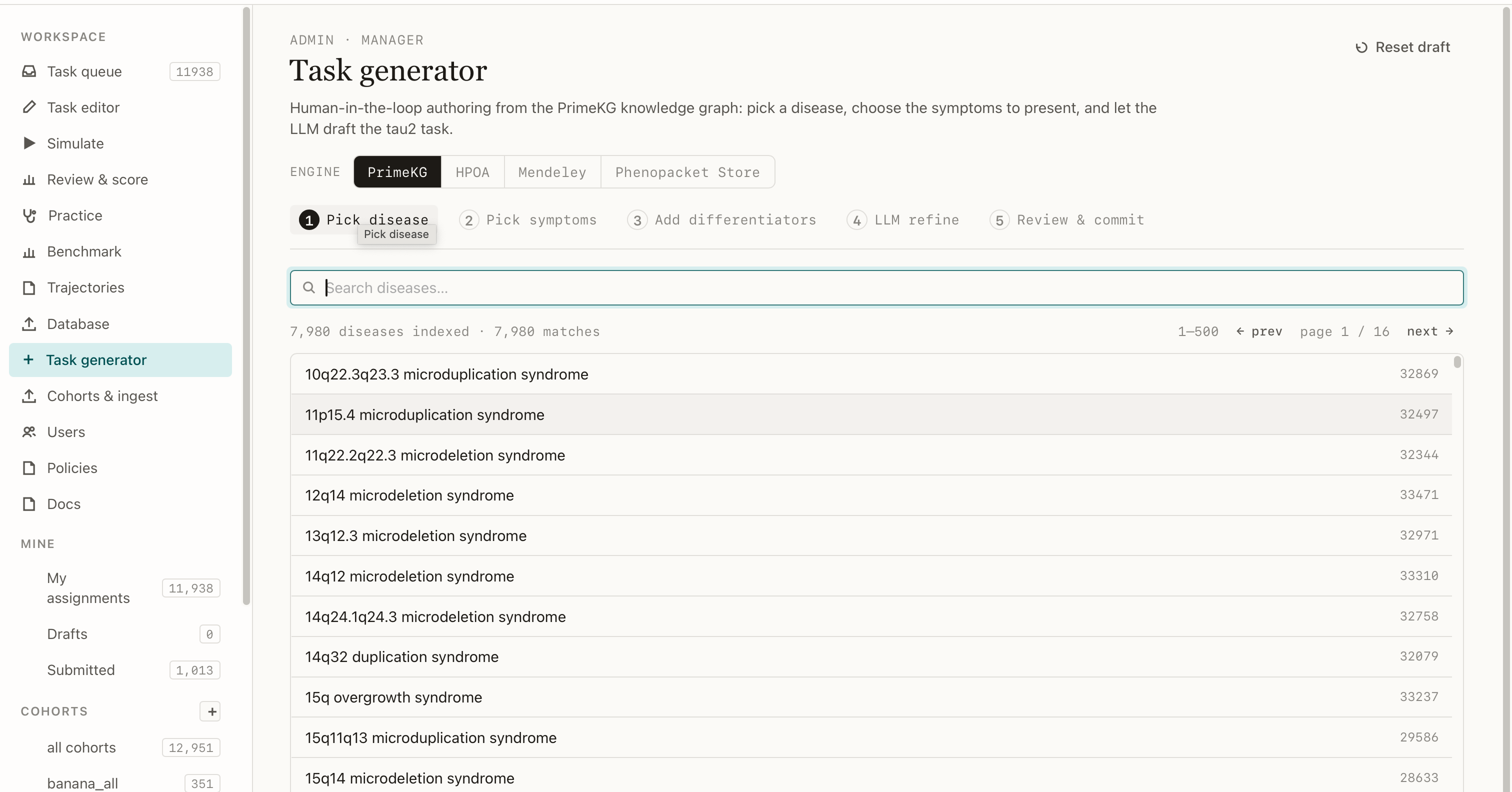}
\caption{Task-authoring interface. The illustrated PrimeKG view supports disease search and a guided sequence of disease selection, symptom selection, differentiating findings, LLM refinement, and clinician review. Workspace counters describe the authoring workspace.}
\label{fig:platform-authoring}\label{fig:generator}
\end{figure}

The task generator starts from a target disease and associated findings (\Cref{fig:platform-authoring}). In the illustrated PrimeKG view, clinicians can search the disease index and select a condition. The interface then guides them through symptom selection, the addition of differentiating findings, LLM refinement, and review of the resulting draft. These interface steps implement the first stage of the construction process in \Cref{sec:platform-generator}: working backwards from a target diagnosis to a patient presentation that an agent must investigate.

Clinicians write the patient's background, history, communication style, and disclosure instructions, together with the accepted diagnosis, required clinical actions, and must-ask topics. LLM assistance helps refine the narrative and translate the authored content into the task schema. Clinicians review and revise the draft before committing it. The task queue and editor provide access to cases as they move through authoring and review; the released benchmark consists of 333 clinician-approved tasks.

\subsection{Live simulation, inspection, and revision}\label{app:live-simulation}

\begin{figure}[H]
\centering
\includegraphics[width=\linewidth]{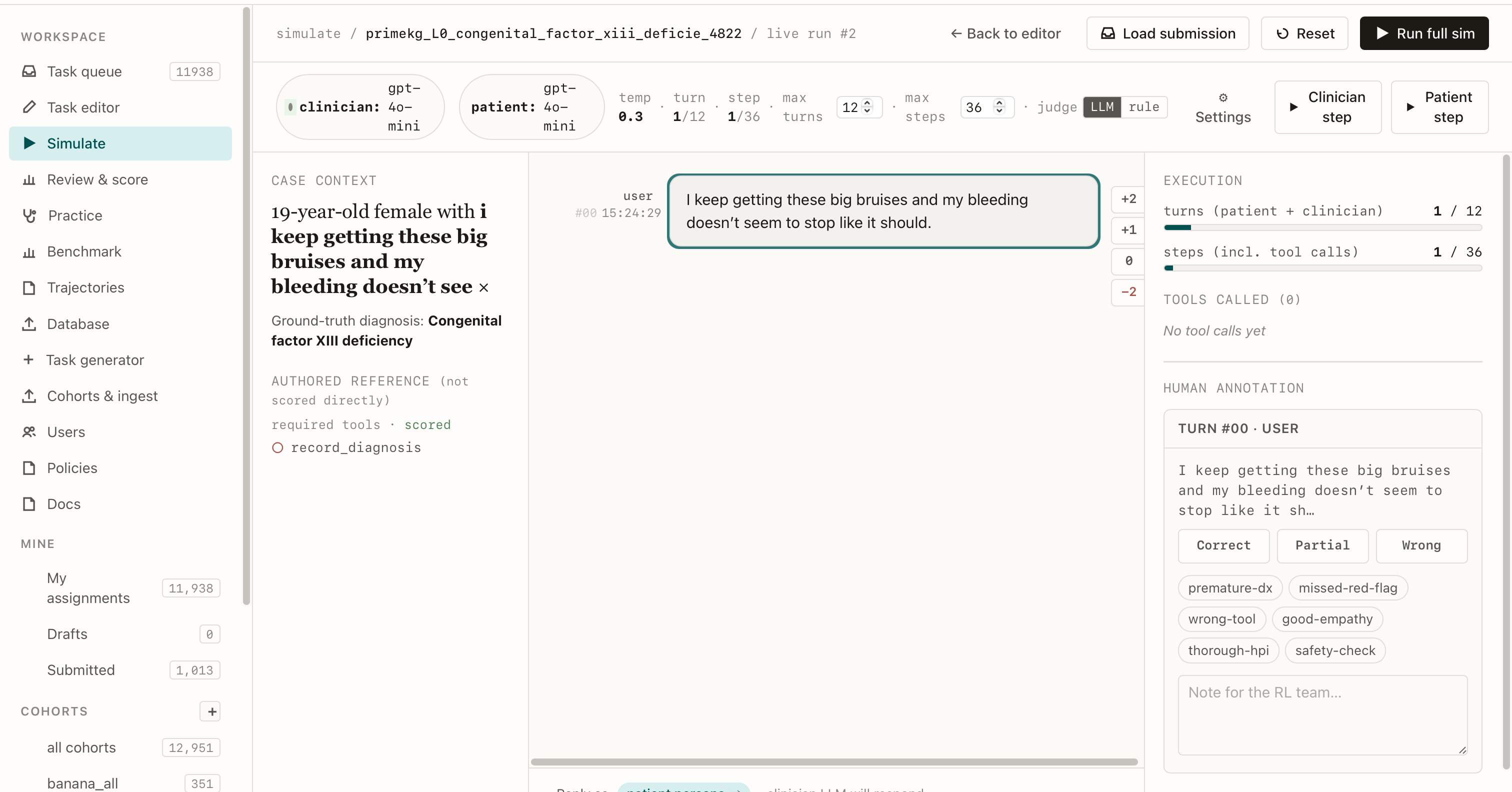}
\caption{Live-simulation and review interface. Clinicians inspect the case context, dialogue, execution counters, and tool calls, and annotate individual turns before returning to the editor. The displayed settings belong to this authoring session; benchmark evaluation follows \Cref{sec:eval}.}
\label{fig:platform-simulation}\label{fig:workflow}
\end{figure}

The simulation workspace lets clinicians test how an authored persona behaves during an encounter (\Cref{fig:platform-simulation}). It places the case context and reference requirements beside the patient--agent dialogue, with execution counters and a record of tool calls on the right. The reference diagnosis is visible to the clinician reviewing the case; the evaluated agent accesses the clinical briefing and tools, while the scoring key remains hidden. Authors can select the clinician-agent and patient models, adjust the simulation settings, advance either participant one step at a time, or run the full encounter.

The human-annotation panel supports turn-level judgments, including \emph{Correct}, \emph{Partial}, and \emph{Wrong}, together with qualitative tags and free-text notes. Clinicians use the dialogue and tool trace to identify inconsistent patient responses, inappropriate disclosure, missing clinical information, or task requirements that need revision. They then return to the editor, revise the persona or criteria, and repeat the simulation. These annotations support case development; benchmark scores are computed separately using the metrics in \Cref{sec:platform-criteria-metrics}.

After clinician approval, the task specification supplies the patient persona, clinical tools, and hidden evaluation criteria used by KlinikeBench. The approved cases are packaged as Harbor tasks for model evaluation, as described in \Cref{sec:eval}. This separates iterative case development from the fixed task specifications used in model comparisons.

\subsection{Released corpus statistics}\label{app:corpus-statistics}

\Cref{tab:dataset-statistics} summarizes the released cases and their history-taking and tool requirements.

\begin{table}[!ht]
\centering
\caption{Corpus totals and per-task statistics.}
\label{tab:dataset-statistics}
% Editable Overleaf table. Values from figure/dataset_statistics.json.
\begingroup
\small
\setlength{\tabcolsep}{4pt}
\renewcommand{\arraystretch}{1.10}
\begin{tabularx}{\linewidth}{@{}>{\raggedright\arraybackslash}Xr@{\hspace{14pt}}>{\raggedright\arraybackslash}Xrrr@{}}
\toprule
\textbf{Corpus} & \textbf{Total} & \textbf{Per task} & \textbf{Mean} & \textbf{Median} & \textbf{Min--Max} \\
\midrule
Clinical tasks & 333 & Must-ask topics & 4.5 & 4 & 0--26 \\
Distinct tool names & 140 & Required tools & 5.8 & 5 & 1--19 \\
Clinical subcategories & 17 & Exposed tools & 10.7 & 10 & 3--24 \\
\bottomrule
\end{tabularx}
\endgroup

\end{table}

\clearpage
\section{Clinical tools}\label{app:clinical-tools}

KlinikeBench exposes a task-specific subset of 140 tool names through the Harbor environment described in \Cref{sec:eval}. Each task provides between 3 and 24 tools (mean 10.7). The catalogue comprises 21 built-in tools, 12 aliases of built-in tools, six additional state-update actions, and 101 named diagnostic studies. \Cref{tab:clinical-tool-catalogue} summarizes their functions; \Cref{fig:corpus-stats} reports their frequency in the task definitions.

\subsection{Interface and task-specific availability}\label{app:tool-interface}

The agent discovers tools through \texttt{clinic tools}, which returns their names, descriptions, and JSON parameter schemas. Most tools require the active \texttt{patient\_id}; additional fields depend on the action. For example, a laboratory order takes \texttt{test\_name}, an imaging order takes \texttt{imaging\_type} and \texttt{body\_region}, and a prescription takes \texttt{drug\_name} and \texttt{dose}, with optional route and frequency. Task-specific tools use a generic schema accepting a patient identifier and additional descriptive arguments. The following example illustrates the command interface; the identifier must be replaced with that of the current patient.

\begin{lstlisting}[basicstyle=\ttfamily\small]
clinic tools
clinic say "When did your symptoms begin?"
clinic call order_lab_test '{"patient_id":"P1","test_name":"CBC"}'
clinic transcript
\end{lstlisting}

Tool availability is derived from each task's required and optional lists. The diagnosis and differential-recording tools are included by default, while forbidden tool names are excluded. The daemon rejects calls to tools outside the exposed set and arguments that are not JSON objects. Each accepted call and its response are appended to the encounter transcript. \texttt{record\_diagnosis} stores the final diagnosis, confidence, and reasoning, and ends the encounter; \texttt{record\_differential} records candidate diagnoses without ending it.

\begin{table}[!ht]
\centering
\caption{Clinical tool catalogue. Counts refer to distinct exposed names across the corpus. The first six rows partition the 21 built-in tools; the final three describe the 119 additional names. Examples are representative, not exhaustive.}
\label{tab:clinical-tool-catalogue}
\small
\setlength{\tabcolsep}{4pt}
\renewcommand{\arraystretch}{1.14}
\begin{tabularx}{\linewidth}{@{}p{2.4cm}r>{\raggedright\arraybackslash}X>{\raggedright\arraybackslash}X@{}}
\toprule
Function & Names & Example tools & Execution behavior \\
\midrule
Patient records & 6 & \texttt{get\_patient\_info}, \texttt{get\_patient\_history}, \texttt{get\_patient\_vitals} & Return demographics, history, medications, allergies, family history, or vital signs. \\
Tests and imaging & 3 & \texttt{order\_lab\_test}, \texttt{get\_lab\_results}, \texttt{order\_imaging} & Store orders and retrieve available laboratory results. \\
Treatment and referral & 3 & \texttt{prescribe\_medication}, \texttt{refer\_to\_specialist}, \texttt{check\_drug\_interactions} & Store prescriptions or referrals; return an interaction-check response. \\
Diagnosis recording & 2 & \texttt{record\_diagnosis}, \texttt{record\_differential} & Store the final diagnosis or differential; only the final diagnosis ends the encounter. \\
Clinical knowledge & 4 & \texttt{get\_treatment\_guidance}, \texttt{get\_disease\_symptoms} & Expose knowledge-query interfaces; return unavailable status in the current headless runtime. \\
Encounter memory & 3 & \texttt{take\_note}, \texttt{set\_working\_diagnosis}, \texttt{summarize\_history\_taken} & Store notes and interim hypotheses, or summarize questions already asked. \\
\midrule
Aliases & 12 & \texttt{get\_patient\_labs}, \texttt{get\_vital\_signs} & Delegate to the corresponding built-in handler. \\
Additional state actions & 6 & \texttt{create\_follow\_up\_plan}, \texttt{admit\_to\_hospital} & Append structured plans, admissions, or referrals to encounter state. \\
Named diagnostic studies & 101 & \texttt{Electromyography\_EMG} & Return cached findings for the current task and study. \\
\bottomrule
\end{tabularx}
\end{table}

\subsection{State updates and returned findings}\label{app:tool-state-updates}

\paragraph{Records and clinical actions.} Read tools consult the available clinical database or the task's patient persona and initial state. Write tools update the simulated encounter: laboratory and imaging orders receive identifiers, while prescriptions, referrals, admissions, and plans are stored as structured records. These actions do not trigger external clinical services. In particular, a generic test order acknowledges an order; it does not itself generate a new measurement.

\paragraph{Case-specific studies.} In the findings-enabled build, named studies retrieve precomputed responses indexed by task identifier and tool name. The cache contains 132 task--study entries spanning 101 tool names: 130 provide results and two mark a study as not indicated. Entries can include narrative findings, measurements, units, and reference ranges. They are authored offline by an LLM and checked for explicit diagnosis leakage before packaging. Tool execution reads the fixed cache without generating findings online. These responses are simulated clinical observations, not measurements from real patients.

\paragraph{Missing information and defaults.} Response behavior depends on the handler. A requested laboratory result that is absent may return \texttt{not\_available}; generic retrieval of ordered tests can instead return a result explicitly marked as mocked. Missing vital signs use fixed fallback values, and some history, medication, or allergy fields use default entries. The drug-interaction tool returns a mock database response. Knowledge-graph query tools and their aliases return \texttt{not\_available} in the headless runtime. The implementation therefore distinguishes executable tool coverage from the availability of case-specific clinical evidence.

\newpage
\subsection{Execution traces and scoring}\label{app:tool-traces}

The runtime records tool names, JSON arguments, and returned responses alongside patient dialogue. The required-tools gate checks whether every required name appears in this trace; it does not independently verify the clinical usefulness of the returned data. Argument-level checks apply the task's configured action constraints. Aliases execute the corresponding handler but retain the invoked name in the trace, so agents must use the tool names exposed for their task. Final diagnosis and must-ask coverage are evaluated separately, as defined in \Cref{sec:platform-criteria-metrics}.

A local dispatch audit of the findings-enabled corpus covered all 3,550 exposed task--tool pairs: 3,003 used built-in handlers, 37 used aliases, 378 used additional state actions, and 132 returned cached study entries. No pair reached the generic echo fallback or raised a handler exception under the representative audit arguments. This check establishes handler coverage for the released catalogue; it does not validate every possible argument or the clinical accuracy of every response.

\clearpage
\section{Training details}\label{app:training-details}

We train Qwen3-4B~\citep{qwen3} as the clinician using the same Harbor task format and \texttt{Terminus-2} harness as the frontier-model evaluation. The findings-enabled corpus is partitioned into 283 training tasks and 50 held-out test tasks (\Cref{sec:data-split}). The test set is a fixed random sample selected with seed 0 and is also used for the ablations. Of its 50 tasks, 42 have a diagnosis that does not occur in the training set. \Cref{tab:training-setup} summarizes the optimization settings.

\paragraph{Supervised fine-tuning.} We collect trajectories from 27 frontier-model configurations on the training tasks, with GPT-5.4-mini as both patient simulator and judge. We retain only trajectories passing all four gates: correct diagnosis, all required tools, full must-ask coverage, and parameterised actions. This yields 1,249 trajectories covering 187 training tasks; removing sequences longer than 20,480 tokens leaves 1,235 trajectories. Each trajectory reconstructs the agent's original multi-turn chat, including the task prompt, assistant JSON replies containing analysis, plans, and shell commands, and the terminal output after each turn. Every assistant reply is prefixed with an empty thinking block, and the loss is computed only on assistant turns. We fine-tune all model parameters for three epochs using FSDP on eight A100 GPUs, with FP32 master weights and BF16 autocast.

\begin{table}[!ht]
\centering
\caption{Training settings for Qwen3-4B. The two RL runs share these settings and differ only in the optimized reward. The SFT patient and judge generate and score the demonstration trajectories.}
\label{tab:training-setup}
\small
\setlength{\tabcolsep}{5pt}
\renewcommand{\arraystretch}{1.08}
\begin{tabularx}{\linewidth}{@{}l>{\raggedright\arraybackslash}X>{\raggedright\arraybackslash}X@{}}
\toprule
Setting & SFT & RL (GRPO) \\
\midrule
Initialization & Qwen3-4B & SFT checkpoint \\
Learning rate & $10^{-5}$ & $10^{-6}$ \\
Schedule & Cosine; 5\% warm-up & Constant \\
Batch & 32 sequences (global) & 8 tasks $\times$ 8 rollouts \\
Training duration & 3 epochs & 315 steps planned; results through 30 \\
Sequence limit & 20,480 tokens & 32k tokens \\
Patient and judge & GPT-5.4-mini & Claude Opus 4.6 \\
Rollout temperature & --- & 1.0 \\
Rollout limits & --- & 32 agent turns; 15 minutes \\
Hardware & 8 A100 GPUs & 4 A100 GPUs per run \\
Execution & FSDP & Shared vLLM generation / FSDP training \\
\bottomrule
\end{tabularx}
\end{table}

\paragraph{Reinforcement learning.} Starting from the SFT checkpoint, we apply GRPO~\citep{grpo} using SkyRL~\citep{skyrl}. We do not normalize advantages by the group's standard deviation and use no KL penalty. The loss averages sequence-level sums of token contributions. Each step samples eight training tasks and generates eight rollouts per task at temperature 1.0. Rollouts are capped at 32 agent turns and 15 minutes; we train on the full trajectory up to 32k tokens and filter out overlong rollouts. Generation with vLLM and training with FSDP share four A100 GPUs per run. Claude Opus 4.6 serves as both the patient simulator and must-ask judge during RL.

\paragraph{Reward comparison.} Using the gate definitions in \Cref{sec:platform-criteria-metrics}, we compare a diagnosis-only reward with a weighted multi-gate reward:
\begin{equation}
 r_i^{\mathrm{diag}} = D_i,
 \qquad
 r_i^{\mathrm{multi}} = \frac{2D_i + W_i + A_i + H_i}{5}.
 \label{eq:training-rewards}
\end{equation}
Both rewards lie in $[0,1]$. The multi-gate reward uses fractional must-ask coverage $H_i$, providing credit for partial history-taking. The two runs share the same SFT initialization, data order, and hyperparameters, with 315 steps (nine epochs) planned per run. The reported results cover the first 30 steps. Every rollout logs all gates regardless of the reward being optimized. We evaluate each test task with one rollout before RL training, at steps 5, 10, and 15, and at subsequent recorded checkpoints. The reported early checkpoints include steps 20 and 30. Base and SFT summaries pool the two initial evaluation passes, giving two rollouts per task; RL checkpoints use one rollout per task.

\clearpage
\section{Additional result details}\label{app:additional-results}

\subsection{Complete model results}\label{app:complete-model-results}

\begin{table}[!ht]
\centering
\caption{Results (\%) for 31 models on 333 tasks (metrics: \Cref{sec:platform-criteria-metrics}). Strict \texttt{pass@1} requires correct diagnosis $\land$ complete clinical actions $\land$ full must-ask coverage. \textbf{Bold}: overall best; \underline{underlining}: best within families with multiple models.}
\label{tab:gates-results-full}
\small
\renewcommand{\arraystretch}{0.90}
\setlength{\aboverulesep}{1.5pt}
\setlength{\belowrulesep}{1.5pt}
\setlength{\tabcolsep}{2pt}
\begin{tabular*}{\linewidth}{@{\extracolsep{\fill}}lrrrr@{}}
\toprule
Model & Strict \texttt{pass@1} & Diagnosis & Tools & Must-ask \\
\midrule
Claude Opus 5~\citep{claudeopus5} & \textbf{\underline{29.4\%}} & \textbf{\underline{90.7\%}} & 44.1\% & \underline{89.9\%} \\
Claude Opus 5.5~\citep{claudeopus55} & 28.8\% & \textbf{\underline{90.7\%}} & \underline{49.8\%} & 84.4\% \\
Claude Opus 4.8~\citep{claudeopus48} & 16.8\% & 89.2\% & 27.9\% & 85.5\% \\
Claude Sonnet 5~\citep{claudesonnet5} & 13.8\% & 87.4\% & 24.0\% & 85.1\% \\
Claude Sonnet 4.5~\citep{claudesonnet45} & 9.9\% & 82.3\% & 25.8\% & 76.7\% \\
Claude Sonnet 4.6~\citep{claudesonnet46} & 8.4\% & 83.8\% & 22.8\% & 73.8\% \\
Claude Haiku 4.5~\citep{claudehaiku45} & 2.7\% & 72.1\% & 8.1\% & 70.6\% \\
\midrule
GPT-6-astra~\citep{gpt6astra} & \underline{24.3\%} & \underline{87.4\%} & 39.3\% & 88.0\% \\
GPT-5.6-sol~\citep{gpt56} & 23.7\% & 86.2\% & 36.6\% & \textbf{\underline{90.8\%}} \\
GPT-5.4~\citep{gpt54} & 23.4\% & 80.2\% & \underline{53.2\%} & 73.5\% \\
GPT-5.5~\citep{gpt55} & 16.8\% & 86.2\% & 30.0\% & 87.5\% \\
GPT-5.4-mini~\citep{gpt54mini} & 15.9\% & 73.9\% & 35.1\% & 80.6\% \\
GPT-5.1~\citep{gpt51} & 15.3\% & 73.9\% & 25.2\% & 89.6\% \\
GPT-5.6-terra~\citep{gpt56} & 8.4\% & 84.7\% & 19.5\% & 85.6\% \\
GPT-5.6-luna~\citep{gpt56} & 7.5\% & 81.1\% & 16.2\% & 84.4\% \\
GPT-4.1 & 7.5\% & 74.2\% & 12.6\% & 89.1\% \\
GPT-4o~\citep{gpt4o} & 3.9\% & 66.4\% & 12.9\% & 79.5\% \\
GPT-5.3-codex~\citep{gpt53codex} & 3.6\% & 84.4\% & 8.7\% & 78.3\% \\
GPT-4.1-mini & 2.1\% & 71.2\% & 6.9\% & 83.1\% \\
GPT-4.1-nano & 0.9\% & 55.6\% & 9.0\% & 56.9\% \\
GPT-4o-mini & 0.6\% & 51.4\% & 6.6\% & 68.6\% \\
\midrule
Gemini-3.8-flash~\citep{gemini38flash} & \underline{18.9\%} & \underline{80.8\%} & \textbf{\underline{53.8\%}} & 76.5\% \\
Gemini-3.5-flash~\citep{gemini35flash} & 15.6\% & 75.7\% & 43.8\% & \underline{76.6\%} \\
Gemini-3.7-flash~\citep{gemini37flash} & 10.8\% & 79.9\% & 53.5\% & 67.1\% \\
\midrule
Grok-4.7~\citep{grok47} & \underline{22.2\%} & \underline{83.2\%} & 35.1\% & 87.5\% \\
Grok-4.6~\citep{grok46} & 21.9\% & 81.4\% & 36.9\% & \underline{89.0\%} \\
Grok-4.3 & 21.3\% & 77.8\% & \underline{45.6\%} & 83.3\% \\
\midrule
DeepSeek-v4-pro~\citep{deepseekv4} & \underline{8.1\%} & \underline{29.4\%} & \underline{13.8\%} & \underline{87.8\%} \\
DeepSeek-v4-flash~\citep{deepseekv4} & 3.9\% & 18.0\% & 9.3\% & 81.3\% \\
\midrule
Kimi-K2.6~\citep{kimik26} & 23.4\% & 77.2\% & 48.7\% & 84.7\% \\
\midrule
GLM-5.1~\citep{glm5} & 4.8\% & 24.9\% & 10.8\% & 30.8\% \\
\bottomrule
\end{tabular*}
\end{table}

Model citations point to verified reports or model cards. For GLM-5.1, we cite the GLM-5 technical report recommended by its \href{https://huggingface.co/zai-org/GLM-5.1}{official model card}; Kimi-K2.6 is cited through its version-specific model card.

\clearpage
\subsection{Cost and performance}\label{app:cost-performance}

\begin{figure}[!ht]
\centering
\includegraphics[width=\linewidth]{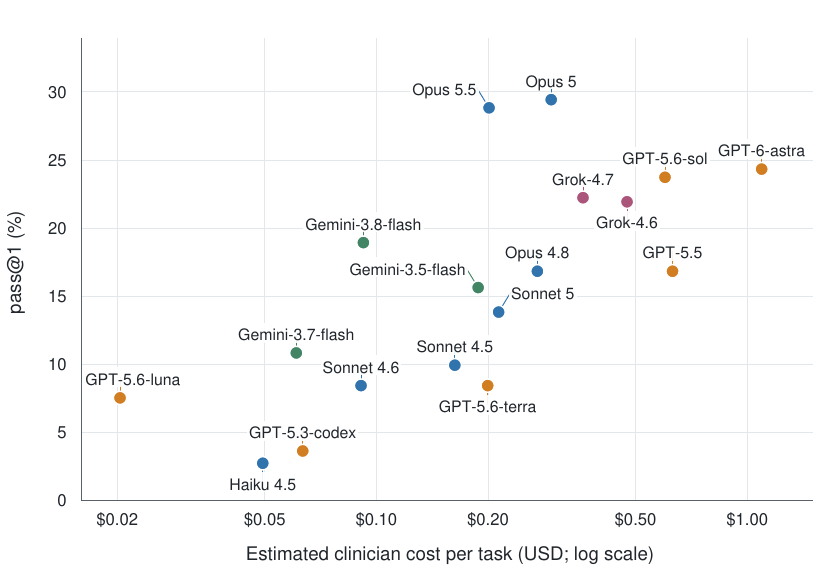}
\caption{Strict \texttt{pass@1} versus estimated clinician cost per task for the 18 models with pricing data. Costs use the workbook's average input, cached-input, and output tokens and listed token rates; patient and judge costs are excluded. The cost axis is logarithmic. Colors indicate model families.}
\label{fig:price-pass-rates}
\end{figure}

\paragraph{Higher inference cost does not guarantee higher success.}
Among models with pricing data, Claude Opus 5 achieves the highest success rate (29.4\%) at an estimated \$0.30 per task, while Opus 5.5 reaches 28.8\% at \$0.20 (\Cref{fig:price-pass-rates}). GPT-6-astra costs \$1.09 per task with 24.3\% success. Gemini-3.8-flash reaches 18.9\% at \$0.09, while GPT-5.6-luna has the lowest estimated cost (\$0.02) and 7.5\% success. These estimates account for model-specific token consumption and cached-input rates; they measure clinician inference cost rather than the total cost of running the benchmark.

\subsection{Interaction-budget scores}\label{app:interaction-budget-scores}

\begin{table}[!ht]
\centering
\caption{Step-cap ablation for five clinicians on 50 matched tasks. Scores are percentages; $\Delta$ is the change in strict \texttt{pass@1} from the default run, in percentage points, with the reported paired-bootstrap 95\% interval. The ten-patient-turn limit remains fixed.}
\label{tab:budget-ablation}
\small
\setlength{\tabcolsep}{3pt}
\begin{tabular*}{\linewidth}{@{\extracolsep{\fill}}llrrrrl@{}}
\toprule
Model & Step cap & \texttt{pass@1} & Diagnosis & Tools & \shortstack{Must-ask\\coverage} & $\Delta$ [95\% CI] \\
\midrule
GPT-6-astra      & Default & 28 & 86 & 46 & 88.9 & --- \\
                 & 45      & 34 & 88 & 60 & 90.4 & $+6\;[-6,20]$ \\
                 & 60      & 32 & 86 & 56 & 91.2 & $+4\;[-8,16]$ \\
                 & None    & 36 & 88 & 52 & 89.0 & $+8\;[-2,20]$ \\
\midrule
Claude Opus 5    & Default & 34 & 86 & 54 & 89.8 & --- \\
                 & 45      & 36 & 84 & 62 & 90.4 & $+2\;[-10,16]$ \\
                 & 60      & 40 & 88 & 60 & 89.9 & $+6\;[-8,20]$ \\
                 & None    & 38 & 88 & 66 & 89.0 & $+4\;[-10,20]$ \\
\midrule
Grok-4.7         & Default & 22 & 78 & 46 & 85.9 & --- \\
                 & 45      & 50 & 78 & 80 & 90.8 & $+28\;[16,40]$ \\
                 & 60      & 38 & 78 & 72 & 86.2 & $+16\;[6,28]$ \\
                 & None    & 30 & 80 & 66 & 84.5 & $+8\;[-4,20]$ \\
\midrule
Gemini-3.8-flash & Default & 20 & 76 & 66 & 74.8 & --- \\
                 & 45      & 22 & 74 & 68 & 76.1 & $+2\;[-12,16]$ \\
                 & 60      & 14 & 72 & 60 & 71.0 & $-6\;[-20,6]$ \\
                 & None    & 20 & 70 & 70 & 74.8 & $0\;[-12,12]$ \\
\midrule
Claude Sonnet 5  & Default & 20 & 76 & 42 & 87.5 & --- \\
                 & 45      & 26 & 82 & 50 & 84.5 & $+6\;[-8,20]$ \\
                 & 60      & 22 & 82 & 44 & 86.8 & $+2\;[-10,14]$ \\
                 & None    & 26 & 84 & 48 & 87.0 & $+6\;[-6,20]$ \\
\bottomrule
\end{tabular*}
\end{table}

\subsection{Full sensitivity results}\label{app:full-sensitivity-results}

\Cref{tab:full-sensitivity} includes every patient and judge condition in the results workbook. A dash indicates a score not yet reported (queued or running), not zero. Rank-agreement statistics in the main text use conditions with all eight scores. The supplied patient-ranking figure additionally includes GPT-6-astra and Claude Sonnet 5 conditions whose numerical scores are partially unavailable in this export.
\begin{table}[!ht]
\centering
\caption{Strict \texttt{pass@1} (\%) for all patient and judge substitutions. OP5/OP5.5: Claude Opus 5/5.5; S5: Claude Sonnet 5; G6/G5.6: GPT-6-astra/GPT-5.6-sol; GR4.7: Grok-4.7; GE3.8/GE3.7: Gemini-3.8/3.7-flash. Dashes mark unavailable scores.}
\label{tab:full-sensitivity}
\small
\setlength{\tabcolsep}{2pt}
\begin{tabular*}{\linewidth}{@{\extracolsep{\fill}}lrrrrrrrr@{}}
\toprule
Condition & OP5 & G6 & GR4.7 & GE3.8 & OP5.5 & G5.6 & S5 & GE3.7 \\
\midrule
\multicolumn{9}{@{}l}{\textit{Judge changed; patient fixed to GPT-5.4-mini}} \\
GPT-5.4-mini (original) & 24 & 30 & 20 & 20 & 26 & 18 & 8 & 8 \\
GPT-5.4-mini (repeat) & 26 & 28 & 24 & 20 & 26 & 18 & 6 & 6 \\
GPT-5.4 & 24 & 32 & 20 & 20 & 32 & 24 & 10 & 6 \\
Gemini-3.5-flash & 22 & 20 & 18 & 12 & 24 & 14 & 6 & 6 \\
Grok-4.5 & 22 & 28 & 18 & 14 & 28 & 16 & 8 & 6 \\
\midrule
\multicolumn{9}{@{}l}{\textit{Patient changed; judge fixed to GPT-5.4-mini}} \\
GPT-5.4-mini (baseline) & 24 & 30 & 20 & 20 & 26 & 18 & 8 & 8 \\
Gemini-3.5-flash & 32 & 30 & 30 & 18 & 24 & 24 & 16 & 8.5 \\
Grok-4.5 & 28 & 30 & 34 & 16 & 34 & 32 & 14 & 10 \\
GPT-5.4 & 38 & 26 & 30 & 16 & 26 & 28 & 24 & 18 \\
Claude Haiku 4.5 & 34 & 28 & 26 & 18 & 34 & 24 & 16 & 16 \\
Claude Opus 4.6 & 28.6 & 32 & 32 & 24 & 32 & 26 & 16 & 16.3 \\
GPT-5.5 & 34 & 30 & 36 & 16 & 28 & 20 & 20 & 16 \\
Gemini-3.8-flash & 28 & 34 & 32 & 10 & 36.2 & 26 & 18 & 16 \\
Grok-4.7 & 27.7 & 30 & 34 & 14 & 34.8 & 16 & 14 & 15.2 \\
GPT-6-astra & 30 & 28 & 24 & 22 & 30 & 26 & 18 & 10 \\
Claude Sonnet 5 & 32 & 30 & 30 & 6 & 32 & 24 & 18 & 14 \\
\bottomrule
\end{tabular*}
\end{table}

\subsection{Qwen3-4B held-out results}

\begin{table}[!ht]
\centering
\caption{Qwen3-4B on 50 held-out tasks. Base and SFT pool two rollouts per task; each RL checkpoint uses one. RL rows report step 30. All scores are percentages.}
\label{tab:qwen-results}
\small
\setlength{\tabcolsep}{3pt}
\begin{tabular*}{\linewidth}{@{\extracolsep{\fill}}lrrrr@{}}
\toprule
Stage & Strict \texttt{pass@1} & Diagnosis & Required tools & Must-ask \\
\midrule
Base & 0.0\% & 26.0\% & 3.0\% & 39.7\% \\
SFT & 8.0\% & 39.0\% & 33.0\% & 77.5\% \\
SFT + RL (diagnosis) & 10.0\% & 44.0\% & 32.0\% & 76.8\% \\
SFT + RL (multi-gate) & 6.0\% & 42.0\% & 32.0\% & 73.1\% \\
\bottomrule
\end{tabular*}
\end{table}

\paragraph{Uncertainty in strict success.}\label{app:qwen-heldout-results}
Paired-bootstrap 95\% confidence intervals describe uncertainty in the change in strict \texttt{pass@1}, measured in percentage points. SFT improves over the base model by 8 points, with an interval from 2 to 15 points. At RL step 30, diagnosis-only RL changes the SFT score by $+2$ points, with an interval from $-7$ to $+12$ points; multi-gate RL changes it by $-2$ points, with an interval from $-11$ to $+7$ points. Negative values indicate worse performance than SFT and positive values indicate better performance. Both RL intervals span zero, so neither comparison establishes an improvement over SFT.

\subsection{Extended GRPO gate trajectories}\label{app:grpo-gate-trajectories}

\begin{figure}[!ht]
\centering

\includegraphics[width=\linewidth]{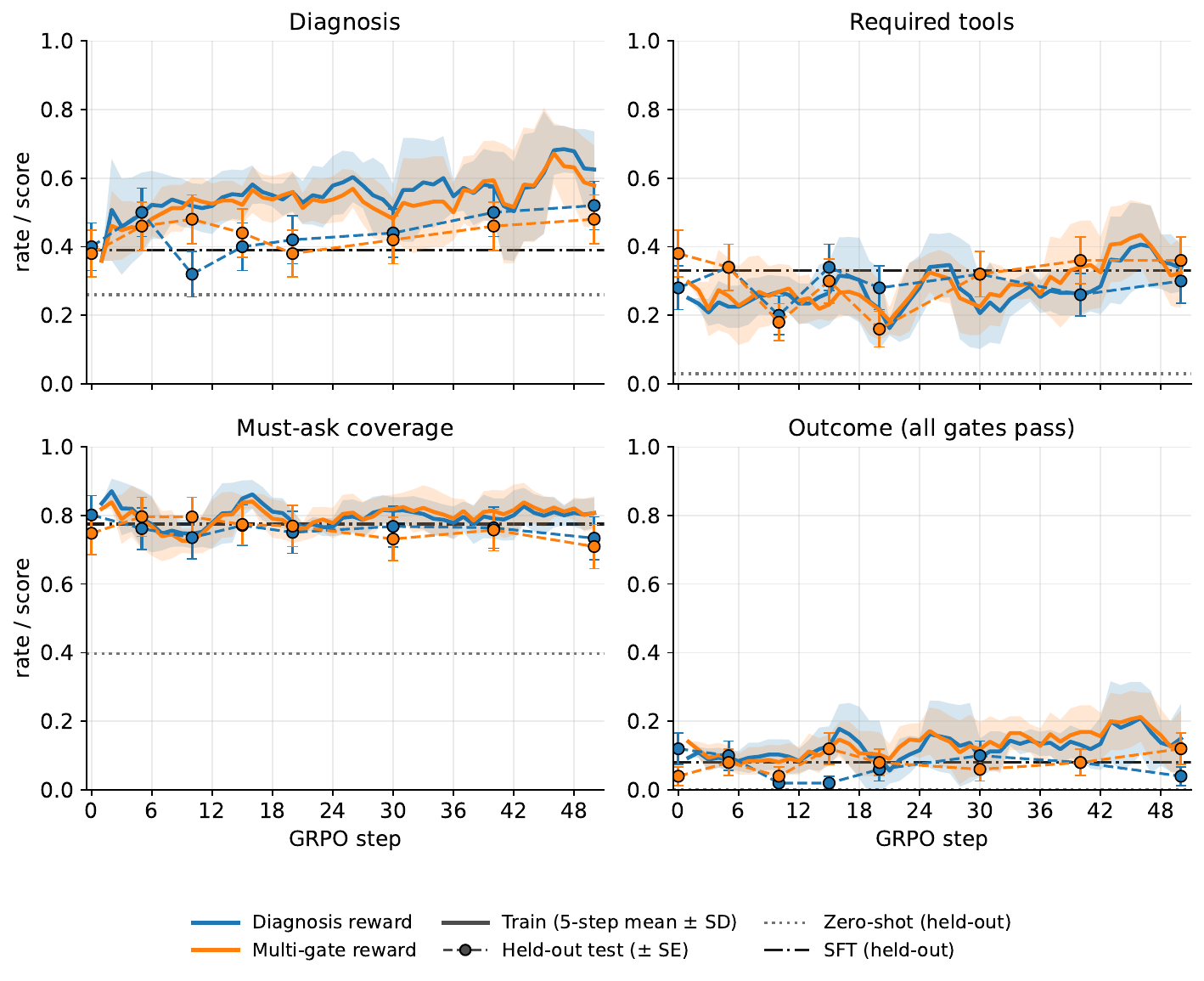}
\caption{Qwen3-4B gate trajectories through GRPO step 50. The top row shows diagnosis and required-tool completion; the bottom row shows must-ask coverage and strict outcome. Blue and orange denote diagnosis-only and multi-gate rewards. Solid curves with shaded bands show five-step training means $\pm$ SD; dashed curves with circles and error bars show held-out results $\pm$ SE. Horizontal dotted and dash-dotted lines mark the zero-shot and SFT held-out baselines. Main-text comparisons use checkpoints through step 30.}
\label{fig:qwen-rl-gates-extended}
\end{figure}

\clearpage
\section{Related work}\label{app:related-work}

\paragraph{Medical knowledge and clinical response quality.}
MedQA, MedMCQA, and PubMedQA evaluate medical knowledge and reasoning from examination questions or biomedical text~\citep{medqa,medmcqa,pubmedqa}, providing a foundation for evaluating clinical language models~\citep{singhal}. MedHELM covers diverse medical tasks, while HealthBench uses physician-written rubrics to assess responses in health conversations~\citep{bedi2026medhelm,health}. RubricsTree structures health-agent evaluation around atomic, clinically verifiable criteria~\citep{zhang2026rubricstree}. KlinikeBench shares the emphasis on explicit clinical criteria, connecting the final diagnosis to the history elicited and actions performed during an encounter.

\paragraph{Interactive diagnostic dialogue.}
AMIE studies diagnostic conversations through simulation-based training and a blinded evaluation with patient actors and primary-care physicians~\citep{patientcoach}. The State Aware Patient Simulator supports automated evaluation of multi-turn consultations~\citep{liao2024saps}. AgentClinic combines simulated patient dialogue, multimodal examinations, and tool use, with evaluation extending to patient-centered aspects of interaction~\citep{agentclinic}. KlinikeBench follows this interactive setting and pairs encounters with clinician-authored must-ask topics, required tools, and action constraints. Our clinician assessment separately examines history disclosure, emotional appropriateness, and human-likeness of the simulated dialogue.

\paragraph{Process-aware consultation benchmarks.}
MedConsultBench is especially close to our motivation: it evaluates the consultation cycle from history taking to follow-up, tracks information acquisition using atomic information units, and identifies deficiencies that a correct diagnosis can conceal~\citep{qiao2026medconsultbench}. MedConceal studies elicitation and management of hidden patient concerns under partial observability, with process and outcome evaluation~\citep{han2026medconceal}. KlinikeBench complements these efforts by coupling clinician-authored information requirements with executable tools and parameterized action constraints. We report each component and use their conjunction with diagnostic correctness to define strict \texttt{pass@1}, distinguishing models with similar diagnostic accuracy by their assessment behavior.

\paragraph{Clinical tools and workflow execution.}
MedAgentBench evaluates clinician-written tasks in an electronic health record environment with standardized FHIR interfaces~\citep{medagentbench}. PhysicianBench studies longer clinical workflows with physician-reviewed checkpoints and execution-grounded verification~\citep{2026PhysicianBench}. MedCUA-Bench evaluates screenshot-based interaction with clinical software using task-completion and safety checks~\citep{2026MedCUABench}. KlinikeBench connects tool execution to patient questioning within the same encounter: information elicited through dialogue and results obtained through tools jointly support diagnosis. Case-specific requirements additionally check that required clinical actions are completed with appropriate parameters.

\paragraph{Longitudinal records and extended conversations.}
LongMedBench evaluates reasoning over longitudinal patient records, while ObGynLongBench studies decisions across pregnancy histories and the difficulty of using evidence embedded in electronic health records~\citep{xu2026longmedbench,xiang2026obgynlongbench}. MedMT-Bench evaluates instruction following in extended medical dialogues; MediLongChat and MedLoCoMo examine history-aware and multi-session conversation~\citep{2026MedMTBench,hu2026medilongchat,zhang2026medlocomo}. These settings emphasize retaining and using information across long contexts or successive visits. KlinikeBench focuses on a complementary horizon: actively gathering evidence and completing the required questions, examinations, and decisions within one assessment under an interaction budget.

\paragraph{Tool--user interaction and evaluation reliability.}
Beyond medicine, $\tau$-bench evaluates tool use and policy compliance through simulated-user interaction, and $\tau^2$-Bench requires coordinated actions by agent and user~\citep{yao2025taubench,barres2026taubench}. KlinikeBench similarly evaluates communication and tool execution, using clinician-defined completion criteria. Since automated judgments can depend on the evaluator~\citep{llmjudge}, we examine judge substitutions on fixed trajectories and patient substitutions that generate fresh encounters, testing how these choices affect scores and rankings.

\end{document}